\pdfoutput=1

\documentclass[11pt]{article}

\usepackage[preprint]{acl}

\usepackage{times}
\usepackage{latexsym}
\usepackage{paralist, tabularx}
\usepackage{multicol}
\usepackage{multirow}
\usepackage{amsmath}
\usepackage{booktabs}
\usepackage{subcaption}
\usepackage{soul}
\usepackage{footmisc}
\usepackage{color}
\usepackage{xcolor}
\usepackage{amsfonts,amssymb}

\usepackage{authblk}

\usepackage[T1]{fontenc}

\usepackage[utf8]{inputenc}

\usepackage{microtype}

\usepackage{inconsolata}

\usepackage{graphicx}

\usepackage[noabbrev,capitalize]{cleveref}

\crefname{equation}{Eq.}{equations}
\crefname{line}{line}{lines}
\crefname{section}{\S}{\S\S}
\Crefformat{section}{#2\S#1#3}
\crefrangeformat{section}{\S\S#3#1#4--#5#2#6}

\title{Process-based Self-Rewarding Language Models}

\author{
\textbf{Shimao Zhang$^\clubsuit$\thanks{Work done during his internship at MSRA.}\ \ 
Xiao Liu$^\bigstar$\thanks{Corresponding authors.}\ \ 
Xin Zhang$^\bigstar$\ \ 
Junxiao Liu$^\clubsuit$\ \ 
Zheheng Luo$^\Diamond$} \\
\textbf{Shujian Huang$^\clubsuit$\footnote[2]\ \ \ 
Yeyun Gong$^\bigstar$} \\
$^\clubsuit$National Key Laboratory for Novel Software Technology, Nanjing University \\
$^\Diamond$The University of Manchester \ \ \ \ 
$^\bigstar$Microsoft Research Asia \\
\texttt{smzhang@smail.nju.edu.cn, huangsj@nju.edu.cn,} \\
\texttt{xiao.liu.msrasia@microsoft.com}
}

\begin{document}
\maketitle

\begin{abstract}
Large Language Models have demonstrated outstanding performance across various downstream tasks and have been widely applied in multiple scenarios. Human-annotated preference data is used for training to further improve LLMs' performance, which is constrained by the upper limit of human performance. Therefore, Self-Rewarding method has been proposed, where LLMs generate training data by rewarding their own outputs. However, the existing self-rewarding paradigm is not effective in mathematical reasoning scenarios and may even lead to a decline in performance. In this work, we propose the Process-based Self-Rewarding pipeline for language models, which introduces long-thought reasoning, step-wise LLM-as-a-Judge, and step-wise preference optimization within the self-rewarding paradigm. Our new paradigm successfully enhances the performance of LLMs on multiple mathematical reasoning benchmarks through iterative Process-based Self-Rewarding, demonstrating the immense potential of self-rewarding to achieve LLM reasoning that may surpass human capabilities.
\footnote{Our code and data will be available at: \url{https://github.com/Shimao-Zhang/Process-Self-Rewarding}.}
\end{abstract}

\section{Introduction}
Large language models (LLMs) acquire powerful multi-task language capabilities through pre-training on extensive corpus~\citep{radford2019language,brown2020language}. Additionally, supervised fine-tuning (SFT) can further effectively improve the model's performance on end-tasks. However, it is found that models after SFT are prone to hallucinations~\citep{lai2024step} due to the simultaneous increasing of the probabilities of both preferred and undesirable outputs~\citep{hong2024orpo}.
Therefore, to further enhance the language capabilities of LLMs to align with human-level performance effectively, researchers often utilize human-annotated preference data for training. A representative approach is Reinforcement Learning from Human Feedback (RLHF)~\citep{christiano2017deep}, which utilizes RL algorithms and external reward signals to help LLMs learn specific preferences.

However, most reward signals rely on human annotations or reward models, which is expensive and bottlenecked by human capability and reward model quality. So the Self-Rewarding Language Models paradigm~\citep{yuan2024self} is proposed to overcome the above limitations, which integrates the reward model and the policy model within the same model. In this framework, a single model possesses the ability to both perform the target task and provide reward feedback. The model can execute different tasks based on the scenario and conduct iterative updates. This paradigm is effective in instruction-following scenarios, where the model achieves performance improvement solely through self-rewarding and iterative updates.

Although the self-rewarding algorithm performs well in the instruction-following tasks, it is also demonstrated that LLMs perform poorly on the mathematical domain data based on the existing self-rewarding algorithm. In fact, model's performance may even degrade as the number of iterations increases~\citep{yuan2024self}. We notice two main limitations in the self-rewarding framework: (a) Existing self-rewarding algorithm is not able to provide fine-grained and accurate reward signals for complex reasoning tasks involving long-thought chains; (b) For a complex mathematical solution, it's hard to design the criterion for generating specific scores. It means that assigning scores to complex long-thought multi-step reasoning for LLMs is more challenging than performing pairwise comparisons, with lower consistency and agreement with humans, which is proven by the results in \cref{appendix:solutionscore-vs-stepcompare}.

In this work, we propose the paradigm of \textit{Process-based Self-Rewarding Language Models}, where we introduce the step-wise LLM-as-a-Judge and step-wise preference optimization into the traditional self-rewarding framework. In a nutshell, we enable the LLMs to simultaneously conduct step-by-step complex reasoning and perform LLM-as-a-Judge for individual intermediate steps. For the limitation (a) above, to get finer-grained and more accurate rewards, Process-based Self-Rewarding paradigm allows LLMs to perform step-wise LLM-as-a-Judge for the individual reasoning step. Since producing the correct final answer does not imply that LLMs can generate correct intermediate reasoning steps, it is crucial to train the model to learn not only to produce the correct final answer but also to generate correct intermediate reasoning steps. By using model itself as a reward model to generate step preference pairs data, we further perform step-wise preference optimization. For the limitation (b) above, we design a LLM-as-a-Judge prompt for step-wise pairwise comparison rather than directly assigning scores to the answer for more proper and steadier judgments based on the observations in \cref{appendix:solutionscore-vs-stepcompare}.

We conduct the experiments on models in different parameter sizes (7B and 72B) and test across a wide range of mathematical reasoning benchmarks. Our results show that Process-based Self-Rewarding can effectively enhance the mathematical reasoning capabilities of LLMs, which indicates that LLMs are able to perform effective self-rewarding at the step level. Our models that iteratively trained based on the Process-based Self-Rewarding paradigm demonstrate an increasing trend in both mathematical and LLM-as-a-judge capabilities. These results suggest this framework’s immense potential for achieving intelligence that may surpass human performance.

\section{Background}

\begin{figure*}[tbp]
    \centering    
    \includegraphics[width=1.0\linewidth]{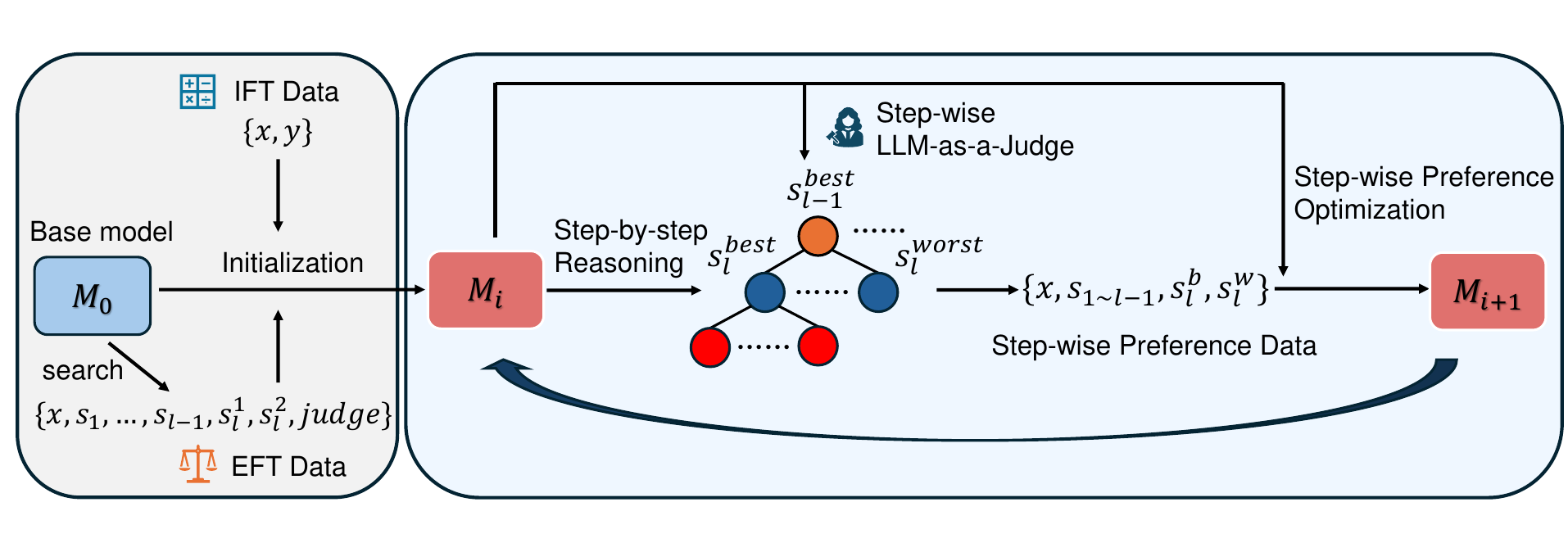}
\caption{\textbf{Illustration of our Process-based Self-Rewarding paradigm.} (1) We get EFT data by tree-search, initial data filtering and data annotation. And we get IFT data by step segmentation. (2) The model is initialized on EFT and IFT data. (3) The model conducts step-by-step search-based reasoning and performs step-wise LLM-as-a-Judge to select the chosen step and generate the step-wise preference pair at each step. (4) We perform step-wise preference optimization on the model. (5) The model enters the next iteration cycle.}
\label{fig:algorithm-pipeline}
\end{figure*}

\subsection{Reinforcement Learning from Human Feedback}\label{sebsec:rlhf}
Supervised Fine-tuning is an effective method to improve LLMs' performance across many different downstream tasks. But it has been evidenced that SFT potentially exacerbates LLMs' hallucination~\citep{hong2024orpo}. So RLHF is further utilized to align LLMs with human preference. In the RLHF paradigm, the model is trained based on reward signals provided by external reward models and humans by reinforcement learning algorithms, such as PPO~\citep{schulman2017proximal}, DPO~\citep{rafailov2024direct}, SimPO~\citep{meng2024simpo}, and so on. Direct Preference Optimization (DPO) is a preference learning algorithm which directly uses pairwise preference data, including chosen and rejected answers for optimization. Furthermore, the step-wise preference optimization has also been investigated for long-chain reasoning and has shown great performance~\citep{lai2024step, chen2024step}. In our work, we introduce the step-wise preference optimization into our Process-based Self-Rewarding paradigm for more fine-grained learning.

\subsection{LLM-as-a-Judge}\label{sebsec:llm-as-a-judge}
LLM-as-a-Judge technique has been widely used for evaluation tasks because of LLMs' scalability, adaptability, and cost-effectiveness~\citep{gu2024survey}. In the LLM-as-a-Judge scenarios, LLMs are prompted to mimic human reasoning and evaluate specific inputs against a set of predefined rules. To improve the performance of LLM-as-a-Judge, the LLM acting as the evaluator is trained to align with human preferences. When conducting LLM-as-a-Judge, LLMs can play many different roles depending on the given prompt. Typical applications include tasks where LLMs are prompted to generate scores~\citep{li2023generative, xiong2024llava}, perform pairwise comparisons~\citep{liu2024aligning, liusie2023zero}, rank multiple candidates~\citep{yuan2023rrhf}, and so on. However, the LLM-as-a-Judge for individual mathematical reasoning steps has not been widely investigated. In our experiment, we design the step-wise LLM-as-a-Judge for rewarding and analyze its performance.

\subsection{Self-Rewarding Language Models}\label{sebsec:srlm}
Although RLHF has been widely utilized to align LLMs with human-level performance and has achieved impressive performance, the existing methods heavily rely on high-quality reward models or human feedback, which bottlenecks these approaches. To avoid this bottleneck, \citet{yuan2024self} propose the Self-Rewarding Language Models paradigm, which uses a single model as both instruction-following model and reward model simultaneously. The iterative self-rewarding algorithm operates by having the model generate responses and reward the generated response candidates, then selecting preference pairs for training. Based on this, \citet{wu2024meta} further improve the judgment agreement by adding the LLM-as-a-Meta-Judge action into the self-rewarding pipeline, which allows the model to evaluate its own judgments. But the existing self-rewarding methods mainly focus on the instruction-following tasks and perform poorly in the mathematical domain data~\citep{yuan2024self}. And evaluating the entire response makes it difficult for the model to learn fine-grained preference information. For some long-thought reasoning tasks, it is important to enable LLMs to focus on and learn the fine-grained reasoning step preference information.

\subsection{Step-by-step Reasoning}\label{sebsec:step-by-step-reasoning}
Complex reasoning tasks are still great challenges for LLMs now. Chain-of-Thought~\citep{wei2022chain} methods prompt LLMs to solve the complex problems by reasoning step by step rather than generating the answer directly, which leads to significant improvements across many reasoning tasks~\citep{yoran2023answering, fu2022complexity, zhang2022automatic}. Furthermore, recent studies investigate the test-time scaling paradigm which allows the LLMs to use more resources and time for inference to achieve better performance~\citep{lightman2023let} typically based on search and step selecting~\citep{yao2024tree, wang2024math}. These results highlight the importance of conducting high-quality long-thought step-by-step reasoning for LLMs in solving complex reasoning problems.

\section{Process-based Self-Rewarding Language Models}
In this section, we propose our new Process-based Self-Rewarding Language Models pipeline. We first review the existing self-rewarding algorithm and our motivation as a preliminary study in \S\ref{subsec:preliminary}. Then we introduce our novel paradigm for more fine-grained step-wise self-rewarding and self-evolution. The entire pipeline consists of sequential stages: model initialization (\S\ref{subsec:initialization}), reasoning and preference data generation (\S\ref{subsec:reasoning-and-data-generation}), and model preference optimization (\S\ref{subsec:model-po}). Finally, we provide a summarized overview of our algorithm (\S\ref{subsec:algorithm-overview}). We illustrate the entire pipeline in Figure \ref{fig:algorithm-pipeline}.

\subsection{Preliminary Study}\label{subsec:preliminary}
Most existing preference optimization algorithms rely on reward signals from external reward models or human-annotated data. However, deploying an external reward model or getting ground truth gold reward signals from human annotators is expensive~\citep{gao2023scaling}. Moreover, due to the inherent limitations and implicit biases of both humans and reward models, these model optimization strategies are bottlenecked~\citep{lambert2024rewardbench, yuan2024self}. Thus, Self-Rewarding algorithm is proposed to mitigate this limitation by enabling the model to provide reward signals for its own outputs and perform self-improvement, showing the feasibility of achieving models that surpass human performance~\citep{yuan2024self}.

There are still many aspects waiting for further research and improvement in the self-rewarding framework. The original method is primarily designed for instruction-following tasks and performs poorly on mathematical reasoning data. Step-by-step long-chain reasoning is widely used for complex mathematical reasoning, which allows the models to conduct more detailed thinking and fine-grained verification of the reasoning steps~\citep{lightman2023let, wang2024math, lai2024step}. Given the effectiveness of step-by-step reasoning, we further propose Process-based Self-Rewarding, introducing LLM-as-a-judge and preference optimization for individual steps.

\subsection{Model Initialization}\label{subsec:initialization}
To perform Process-based Self-Rewarding, models need to possess two key abilities:

\begin{itemize}
    \item Step-by-step mathematical reasoning: When faced with a complex reasoning problem, the model needs to think and reason step by step, outputting the reasoning process in a specified format. (Each step is prefixed with ``Step n: '', where n indeicates the step number.)
    \item LLM-as-a-Judge for individual steps: The model should be able to assess the quality of the given next reasoning steps based on the existing problem and partial reasoning steps and provide a detailed explanation.
\end{itemize}

We construct data separately for the two tasks to perform cold start. Following \citet{yuan2024self}, we refer to them as Instruction Fine-Tuning (IFT) data and Evaluation Fine-Tuning (EFT) data. For IFT data, we divide the given solution steps into individual steps logically without altering any information in the original solution by using OpenAI o1~\citep{jaech2024openai}.

For EFT data, since there is no available step-wise LLM-as-a-Judge dataset, we first train Qwen2.5-72B~\citep{yang2024qwen2} on PRM800k~\citep{lightman2023let} following \citet{wang2024openr}. After getting a Process Reward Model~(PRM) by this, which can output a single label ``+'' or ``-'' for a reasoning step based on the question and the previous steps, we conduct Monte Carlo Tree Search~(MCTS) on a policy model. We use the probability of label ``+'' of the above PRM to compare the relative quality of all candidate steps at the same layer, and choose the best and the worst step as a data pair. After the initial data filtering process, we use GPT-o1 to generate judgments and detailed explanations for the obtained data pairs. The pairs whose judgments align with the previous PRM assessments are selected as the final EFT data. Additionally, to enhance consistency, we evaluate each pair twice using GPT with different input orders and select only the pairs that have consistent results.

\subsection{Step-by-step Long-chain Reasoning and Preference Data Generation}\label{subsec:reasoning-and-data-generation}
After the ``EFT + IFT'' initialization stage, the model is able to conduct both step-wise LLM-as-a-Judge and step-by-step mathematical reasoning in the specified formats. Because we conduct pairwise comparison rather than single answer grading, we utilize the following search strategy:
\begin{equation}
\small
    S_l = \{s_{l,1},\ s_{l,2},\ s_{l,3},\ ...,\ s_{l,w-1},\ s_{l,w}\}
\end{equation}
where $S_l$ is all candidates for the next step, $l$ is the step number starting from $1$, $w$ is a hyperparameter to specify the search width for each step.
\begin{equation}
\small
\textrm{Score}_{l,i} = \sum_{1\leq j\leq w,\ j \neq i}\mathbf{O}(s_{l,i},\ s_{l,j} \mid x,\ s_1, s_2, ..., s_{l-1})
\end{equation}
where $l$ is the next step number, $s_{l,i}$ indicates the $i$-th candidate for the next $l$-th step, $x$ is the prompt, and $\mathbf{O}$ is a function that takes $1$ when $s_{l,i}$ is considered better than $s_{l,j}$ and $0$ otherwise.
\begin{equation}
\small
s_l^{\textrm{best}} = S_l[\max(\textrm{Score}_l)]
\end{equation}
\begin{equation}
\small
s_l^{\textrm{worst}} = S_l[\min(\textrm{Score}_l)]
\end{equation}
\begin{equation}
\small
s_l = s_l^{best}
\end{equation}
where $\max(\textrm{Score}_l)$ is the index of the candidate with the highest score and $\min(\textrm{Score}_l)$ corresponds to the lowest score. $s_l$ is the final chosen $l$-th step. $(s_l^{\textrm{best}},\ s_l^{\textrm{worst}})$ will be chosen as a chosen-rejected preference pair.

This process will be repeated continuously until generation is complete. It is important to note that to enhance the effectiveness of preference data, if $\max(\textrm{Score}_l)$ is equal to $\min(\textrm{Score}_l)$, we will discard the existing $s_{l-1}$ and $(s_{l-1}^{\textrm{best}},\ s_{l-1}^{\textrm{worst}})$ and roll back to the previous step.

\subsection{Step-wise Model Preference Optimization}\label{subsec:model-po}
With preference data collected in the Section \ref{subsec:reasoning-and-data-generation}, we conduct preference optimization training on the model. We choose Direct Preference Optimization~(DPO) as the training algorithm~\citep{rafailov2024direct}. The difference is that we conduct a more fine-grained step-wise DPO in our work. The similar method has also been investigated by \citet{lai2024step}. We can calculate the training loss as:
\begin{equation}
\small
    A = \beta \operatorname{log}\frac{\pi_\theta (s_l^{b} \mid x,s_1,...,s_{l-1})}{\pi_{ref} (s_l^{b} \mid x,s_1,...,s_{l-1})}
\end{equation}
\begin{equation}
\small
    B = \beta \operatorname{log}\frac{\pi_\theta (s_l^{w} \mid x,s_1,...,s_{l-1})}{\pi_{ref} (s_l^{w} \mid x,s_1,...,s_{l-1})}
\end{equation}
\begin{equation}
\small
    \mathcal{L}(\pi_\theta;\pi_{ref}) = -\mathbb{E}_{(x,s_1,...,s_l^{b},s_l^{w}) \sim \mathcal{D}}[\operatorname{log}\sigma(A-B)]
\end{equation}
where $x$ is the prompt, $s_1,...,s_{l-1}$ is the previous steps, $s_l^b$ and $s_l^w$ are the best and worst steps respectively for the $l$-th step, $\beta$ is a hyperparameter controlling the deviation from the base reference policy, $\pi_\theta$ and $\pi_{ref}$ are the policies to be optimized and the reference policy respectively.

After the preference optimization stage,  we have the model for the next cycle. In the next iteration, we sequentially repeat the steps in \S\ref{subsec:reasoning-and-data-generation} and \S\ref{subsec:model-po}.

\subsection{Iteration Pipeline}\label{subsec:algorithm-overview}
We show the entire pipeline of our algorithm. Following \citet{yuan2024self}, we refer to the model after $n$ iterations as $M_n$. And we refer to the Pair-wise Preference Data generated by $M_n$ as PPD($M_n$). Then the sequence in our work can be defined as:
\begin{itemize}
    \item $M_0$: The base model.
    \item $M_1$: The model obtained by supervised fine-tuning~(SFT) $M_0$ on ``EFT + IFT'' data.
    \item $M_2$: The model obtained by training $M_1$ on PPD($M_1$) using step-wise DPO.
    \item $\cdots\cdots$
    \item $M_n$: The model obtained by training $M_{n-1}$ on PPD($M_{n-1}$) using step-wise DPO.
\end{itemize}
In summary, we initialize the base model using well-selected step-wise LLM-as-a-Judge data~(EFT) and step-by-step long-thought reasoning data~(IFT). Once the model possesses the corresponding two abilities, we select preference pairs through search and reward signals provided by the model itself, and train the model using step-wise DPO. Then we iterate the model by repeatedly performing the above operations.

\begin{table*}[tbp]
    \centering
    \small
    \begin{tabular}{lcccccccc}
        \toprule
        \textbf{Model} & \textbf{GSM8k} & \textbf{MATH} & \textbf{Gaokao2023En} & \textbf{OlympiadBench} & \textbf{AIME2024} & \textbf{AMC2023} & \textbf{Avg.} \\
        \midrule
        GPT-4o & 92.9 & 76.6 & 67.5 & 43.3 & 10.0 & 47.5 & 56.3 \\
        \midrule
        \midrule
        \textit{\textbf{7B Base Model}} \\
        \midrule
        $M_0$ & 70.1 & 51.7  & 51.2 & 21.3 & 0.0 & 22.5 & 36.1 \\
        SRLM - $M_1$ & 88.2 & 69.0 & 61.6 & 37.6 & 10.0 & 45.0 & 51.9 \\
        \ \ \ \ \ \ \ \ \ \ \ \ \ \ $M_2$ & 87.6 & 69.4 & 63.9 & 37.2 & 3.3 & 40.0 & 50.2 \\
        \ \ \ \ \ \ \ \ \ \ \ \ \ \ $M_3$ & 88.5 & 70.0 & 61.3 & 36.7 & 10.0 & 40.0 & 51.1 \\
        \ \ \ \ \ \ \ \ \ \ \ \ \ \ $M_4$ & 88.3 & 70.2 & 63.9 & 37.6 & 13.3 & 45.0 & 53.1 \\
        PSRLM - $M_1$ & 88.5 & 69.5 & 61.8 & 36.0 & 6.7 & 45.0 & 51.3 \\
        \ \ \ \ \ \ \ \ \ \ \ \ \ \ \ \ $M_2$ & 88.8 & 69.7 & 63.9 & 36.3 & \textbf{16.7} & 47.5 & 53.8 \\
        \ \ \ \ \ \ \ \ \ \ \ \ \ \ \ \ $M_3$ & 88.5 & 72.2 & 64.7 & \textbf{39.9} & 10.0 & 50.0 & 54.2 \\
        \ \ \ \ \ \ \ \ \ \ \ \ \ \ \ \ $M_4$ & \textbf{88.8} & \textbf{73.3} & \textbf{65.2} & 38.7 & 13.3 & \textbf{55.0} & \textbf{55.7} \\
        \midrule
        \midrule
        \textit{\textbf{72B Base Model}} \\
        \midrule
        $M_0$ & 87.5 & 69.7 & 55.3 & 28.9 & 10.0 & 40.0 & 48.6 \\
        SRLM - $M_1$ & 92.9 & 76.4 & 67.3 & 41.8 & 16.7 & 47.5 & 57.1 \\
        \ \ \ \ \ \ \ \ \ \ \ \ \ \ $M_2$ & 92.1 & 76.1 & 66.8 & 42.1 & 20.0 & 55.0 & 58.7 \\
        \ \ \ \ \ \ \ \ \ \ \ \ \ \ $M_3$ & 92.5 & 75.8 & 67.5 & 42.5 & 20.0 & 52.5 & 58.5 \\
        \ \ \ \ \ \ \ \ \ \ \ \ \ \ $M_4$ & 92.8 & 76.1 & 66.2 & 44.0 & 13.3 & 42.5 & 55.8 \\
        PSRLM - $M_1$ & 92.6 & 75.6 & 67.3 & 41.8 & 13.3 & 45.0 & 55.9 \\
        \ \ \ \ \ \ \ \ \ \ \ \ \ \ \ \ $M_2$ & 92.6 & 76.4 & 67.8 & 41.8 & 20.0 & 57.5 & 59.4 \\
        \ \ \ \ \ \ \ \ \ \ \ \ \ \ \ \ $M_3$ & 93.7 & 76.4 & 67.3 & 42.7 & 23.3 & 52.5 & 59.3 \\
        \ \ \ \ \ \ \ \ \ \ \ \ \ \ \ \ $M_4$ & \textbf{93.7} & \textbf{76.6} & \textbf{68.1} & \textbf{44.1} & \textbf{23.3} & \textbf{57.5} & \textbf{60.6} \\
        \bottomrule
    \end{tabular}
    \caption{Accuracy of Process-based Self-Rewarding based on 7B and 72B base models. SRLM is the self-rewarding language model algorithm as the baseline. We bold the best results for each parameter size in each benchmark.}
    \label{tab:math-results}
\end{table*}

\section{Experimental Setup}
We conduct our experiments on models in different parameter sizes and several representative mathematical reasoning benchmarks. In this section, we introduce our experimental settings in detail.

\paragraph{Models} We choose the base model from Qwen2.5-Math series~\citep{yang2024qwen2math} in our experiments, which is one of the most popular open-source LLM series. Specifically, we choose Qwen2.5-Math-7B and Qwen2.5-Math-72B. Additionally, we choose OpenAI GPT-o1~\citep{jaech2024openai} for our initialization data processing (\S\ref{subsec:initialization}).

\paragraph{Datasets} In our experiments, we mainly focus on two capabilities of the model:
\begin{itemize}
    \item \textbf{Step-by-step Mathematical Reasoning: }We choose a subset of NuminaMath~\citep{numina_math_datasets} for IFT data construction, whose solutions have been formatted in a Chain of Thought (CoT) manner. We extract a subset of 28,889 samples and prompt GPT-o1~\citep{jaech2024openai} to logically segment the solutions into step-by-step format without altering any original content. The corresponding prompt is presented in Figure~\ref{fig:construct-steps-prompt}. And the instruction format for step-by-step long-thought reasoning is presented in Figure~\ref{fig:math-prompt}.
    \item \textbf{Step-wise LLM-as-a-Judge: }As described in the Section \ref{subsec:initialization}, we first filtrate some preference pairs using the trained PRM. Then we utilize GPT-o1 and get a total of 4,679 EFT data with judgments and detailed explanations. Finally we split the whole dataset into 4,167 samples as the training set and 500 samples as the test set. The instruction format for step-wise pairwise LLM-as-a-Judge is presented in Figure~\ref{fig:judge-prompt}, which is following the basic format of \citet{zheng2023judging}.
\end{itemize}

And for mathematical task evaluation, following \citet{yang2024qwen2math}, we evaluate the LLMs' mathematical capabilities across some representative benchmarks. We choose the widely used benchmarks GSM8k~\citep{cobbe2021training} and MATH~\citep{hendrycks2021measuring}. We also choose some complex and challenging competition benchmarks, including Gaokao2023En~\citep{liao2024mario}, Olympiadbench~\citep{he2024olympiadbench}, AIME2024\footnote{\url{https://huggingface.co/datasets/AI-MO/aimo-validation-aime}}, and AMC2023\footnote{\url{https://huggingface.co/datasets/AI-MO/aimo-validation-amc}}.

\paragraph{Evaluation Metrics} We use accuracy as the evaluation metric for both the mathematical performance and LLM-as-a-Judge quality. For accuracy calculation on mathematical benchmarks, we follow the implementation of \citet{yang2024qwen2math}.

\paragraph{Implementations} For initial PRM training, we fine-tune full parameters on 128 NVIDIA A100 GPUs for 1 epoch with learning\_rate=$1e-5$ and batch\_size=$128$. For preliminary preference pairs selection, we set simulation\_depth=$3$, num\_iterations=$100$, T=$0.7$, and top\_p=$0.95$. When training $M_0$ to $M_1$, we utilize 28,889 IFT and 4,179 EFT samples. We fine-tune LLMs' full parameters on 32 NVIDIA H100 GPUs for 3 epochs with learning\_rate=$1e-6$ and batch\_size=$32$. During the reasoning and preference data generation stage, we utilize temperature sampling which trade-off generation quality and diversity~\citep{zhang2024edt}. We set T=$0.5$, top\_p=$0.95$. The search width for each step is set to $6$, and the max iteration number is set to $20$. Finally, in the step-wise preference optimization, we train LLM's full parameters on 32 NVIDIA H100 GPUs for 1 epoch with learning\_rate=$5e-7$ and batch\_size=$32$. To get models from $M_2$ to $M_4$, we use $400$, $800$, and $1,200$ math questions for preference pairs generation respectively, which are all sampled from the train subset of NuminaMath. For all solution-scoring judgment strategy experiments, we use the same prompt template of \citet{yuan2024self}. We use greedy search in evaluations.

\begin{table*}[tbp]
    \centering
    \small
    \begin{tabular}{lccccccc}
        \toprule
        \textbf{7B} & \textbf{GSM8k} & \textbf{MATH} & \textbf{Gaokao2023En} & \textbf{OlympiadBench} & \textbf{AIME2024} & \textbf{AMC2023} \\
        \midrule
        SRLM & +0.1 & +1.2 & +2.3 & 0.0 & +3.3 & 0.0 \\
        Process-based (Ours) & +0.3 & +3.8 & +3.4 & +2.7 & +6.6 & +10.0 \\
        \bottomrule
        \toprule
        \textbf{72B} & \textbf{GSM8k} & \textbf{MATH} & \textbf{Gaokao2023En} & \textbf{OlympiadBench} & \textbf{AIME2024} & \textbf{AMC2023} \\
        \midrule
        SRLM & -0.1 & -0.3 & -1.1 & +2.2 & -3.4 & -5.0 \\
        Process-based (Ours) & +1.1 & +1.0 & +0.8 & +2.3 & +10.0 & +12.5 \\
        \bottomrule
    \end{tabular}
    \caption{The results of LLMs' mathematical performance changes after all iterations from $M_1$ to $M_4$.}
    \label{tab:delta-math-results}
\end{table*}

\section{Results}\label{sec:results}
In this section, we report our main results on different mathematical benchmarks and conduct some discussions and analyses based on the results.

\subsection{Main Results}\label{subsec:main-results}
We report the performance of $M_0$ to $M_4$ based on Qwen2.5-Math-7B and Qwen2.5-Math-72B respectively in Table \ref{tab:math-results}. Our findings are as follows:

 \textbf{As the number of iterations increases, the overall performance of the model improves.} Traditionally, external reward signals and training data are utilized for improving LLMs' performance. Our results indicate that models' overall performance on mathematical tasks significantly improves from $M_1$ to $M_4$ solely through Process-based Self-Rewarding and step-wise preference optimization without any additional guidance. This leverages the potential of LLMs for both mathematical reasoning and as evaluators.

\textbf{Our fine-grained algorithm outperforms the tranditional method.} After three iterations, our approach achieves superior performance compared to method that applies rewards and conducts training on the entire response. Given that the initialization with different EFT data lead to different M1 fiducial performance in the two methods, we also report the performance changes from M1 to M4 after multiple iterations in Table \ref{tab:delta-math-results}, which reflects the algorithm's effectiveness and stability in improving the model's mathematical capabilities. Our method achieves more stable and effective improvements across all benchmarks. On one hand, using step-wise preference data enables the model to focus on more fine-grained information; on the other hand, conducting LLM-as-a-Judge on individual steps helps the model more easily detect subtle differences and errors.

\textbf{The models show noticeable improvements on some complex tasks.} For some complex and highly challenging benchmarks, such as MATH, AIME2024, and AMC2023, LLMs' performance show significant improvement. Complex problems require multi-step, long-thought reasoning. Our method effectively leverages the model's existing knowledge to optimize the individual intermediate reasoning steps, achieving favorable results.

\textbf{Our method remains effective across models of different parameter sizes.} We validate our method on both 7B and 72B LLMs to strengthen our conclusions. We find performance improvements across models of different parameter sizes on multiple mathematical tasks through Process-based Self-Rewarding. We also find that the 72B model gains more stable improvements compared to the 7B model, whose mathematical reasoning and LLM-as-a-Judge capabilities are stronger.

Overall, we can find that the models iterating based on the Process-based Self-Rewarding paradigm achieve significant improvements across multiple mathematical tasks, outperforming the traditional self-rewarding method.

\subsection{Further Analysis}\label{subsec:analysis}
Based on the above results, we conduct more analysis and observations of the pipeline.

\begin{table}[tbp]
    \centering
    \small
    \begin{tabular}{lcc}
        \toprule
        \textbf{Model} & \textbf{7B} & \textbf{72B} \\
        \midrule
        $M_0$ (3-shot) & 57.2 & 73.4 \\
        $M_1$ & 92.8 ($\uparrow$) & 95.6 ($\uparrow$) \\
        $M_2$ & 91.6 ($\downarrow$) & 95.8 ($\uparrow$) \\
        $M_3$ & 92.0 ($\uparrow$) & 95.2 ($\downarrow$) \\
        $M_4$ & 92.2 ($\uparrow$) & 95.6 ($\uparrow$) \\
        \bottomrule
    \end{tabular}
    \caption{Judgment accuracy in step-wise LLM-as-a-Judge. We report the results of models with different parameter sizes. Additionally, we use arrows to indicate the changes in accuracy during the iterations.}
    \label{tab:llm-as-a-judge-eval}
\end{table}

\paragraph{Step-wise LLM-as-a-Judge Capability.} We evaluate the LLMs' ability to accurately assess reasoning steps as a reward model during the iterative process. We test the model on the test set including 500 samples (\S \ref{subsec:initialization}). We report the results in Table \ref{tab:llm-as-a-judge-eval}. As shown in the table, LLMs achieve strong reward model performance after initialization with a small amount of EFT data, which indicates the immense potential of LLMs for step-wise LLM-as-a-Judge with CoT reasoning. Additionally, we can observe that, under the same conditions, the larger model exhibits stronger capabilities as a reward model than the smaller one.

Additionally, although we mix EFT data and IFT data for initialization and introduce no additional LLM-as-a-Judge data during subsequent iterations, the LLMs' capabilities to perform LLM-as-a-Judge as a reward model are still good. Furthermore, a consistent pattern is observed across different models where evaluation accuracy initially increases, then decreases, and finally rises again. Based on the analysis above, initially, LLMs gain strong evaluation capabilities through training on EFT data. And there is a temporary decline (but very slight) due to training on mathematical data. Ultimately, as the model's mathematical abilities improve, its ability to evaluate mathematical reasoning steps also increases.

\begin{table}[tbp]
    \centering
    \small
    \begin{tabular}{lcccc}
    \toprule
    \multirow{2}*{\textbf{Iterations}} & \multicolumn{2}{c}{\textbf{Step Num}} & \multicolumn{2}{c}{\textbf{Step Length}} \\
    \cmidrule(lr){2-3}\cmidrule(lr){4-5}
    & GSM8k & MATH & GSM8k & MATH \\
    \midrule
    $M_1$ & 5.89 & 8.41 & 47.79 & 61.00 \\
    $M_2$ & 5.55 & 7.64 & 51.19 & 67.17 \\
    $M_3$ & 5.10 & 6.30 & 57.75 & 80.46 \\
    $M_4$ & 4.87 & 5.54 & 62.86 & 96.63 \\
    \bottomrule
    \end{tabular}
    \caption{Statistics of step number and step length on GSM8k and MATH benchmarks based on 72B models. The full results are reported in Appendix \ref{appendix:step-stat}.}
    \label{tab:72B-gsm8k-math-step-stat}
\end{table}

\begin{figure*}[tbp]
    \centering
    \begin{minipage}[b]{0.45\textwidth}
    \includegraphics[width=1.0\linewidth]{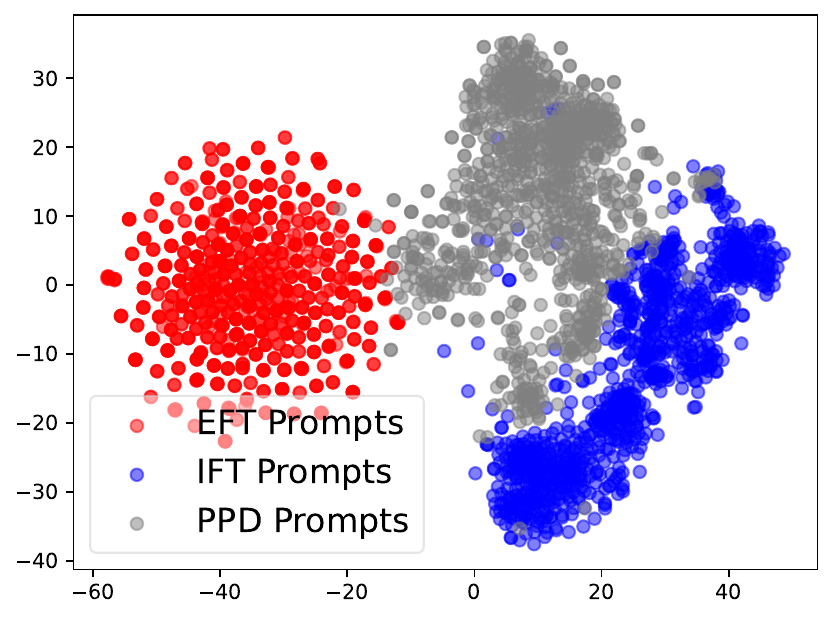}
    \subcaption{Prompt Distributions}
    \label{fig:prompt-distributions}
    \end{minipage}
    \begin{minipage}[b]{0.444\textwidth}
    \includegraphics[width=1.0\linewidth]{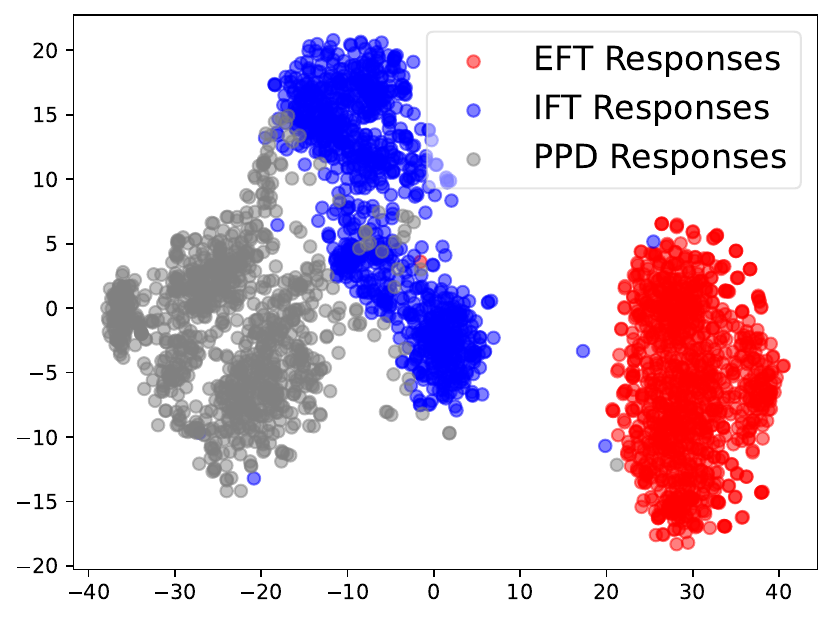}
    \subcaption{Response Distributions}
    \label{fig:response-distributions}
    \end{minipage}
\caption{The data distribution of prompts and responses in EFT (red), IFT (blue) and PPD (grey) data.}
\label{fig:data-distributions}
\end{figure*}

\paragraph{Data Distribution Analysis.} Following \citet{yuan2024self}, we also analyze the distribution of different data. We utilize Bert~\citep{devlin2018bert} for embedding and t-SNE~\citep{van2008visualizing} based on the implementation of \citet{Policar2024} for visualization. We present the results in Figure \ref{fig:data-distributions}. For prompts, the distributions of EFT data and IFT data do not overlap, allowing the model to distinctly learn two different task patterns. For models' responses, we can find the similar phenomenon that the distribution of PPD and IFT responses is distinct from EFT's, which reduces the mutual interference between LLMs' two capabilities during iteration. This allows the model's ability to perform LLM-as-a-Judge to improve alongside its mathematical ability finally, without being overly influenced by the training data itself.

\paragraph{Step Number and Length of Responses.} Step-by-step reasoning is important for LLMs to solve complex reasoning tasks. Therefore, we conduct statistical analysis on the reasoning steps during iterations. As shown in \cref{tab:72B-gsm8k-math-step-stat}, for the same model, more difficult problems require more reasoning steps and longer step lengths. As the iterations progress, the step number across different tasks decreases, while the length of each step increases. This indicates that performing Process-based Self-Rewarding encourages the model to generate longer and higher-quality single reasoning steps, which helps to reach final answers with fewer steps. Additionally, this behavior is also related to LLMs' preferences when performing LLM-as-a-Judge evaluations. More results are in \cref{appendix:step-stat}.

\begin{table}[tbp]
    \centering
    \small
    \begin{tabular}{lcc}
        \toprule
        \textbf{Strategy} & \textbf{Greedy Search} & \textbf{Test-time Scaling} \\
        \midrule
        $M_1$ & 55.9 & 58.2 \\
        $M_4$ & 60.6 & 62.4 \\
        \bottomrule
    \end{tabular}
    \caption{The average results of 72B model on all benchmarks using greedy search or test-time scaling. The full results are reported in Table \ref{tab:test-time-scaling-full}.}
    \label{tab:test-time-scaling}
\end{table}

\paragraph{Test-time Scaling with Process-based Self-Rewarding Language Models.} In the test-time scaling, LLMs conduct step search and select based on the rewards from PRM. Although we don't primarily focus on the test-time scaling performance in our work, LLMs in the Process-based Self-Rewarding paradigm naturally have the ability to perform test-time scaling based on self-rewarding. We perform $6$ generations for each step with the temperature of $0.5$ and select the best one. The results we report in Table \ref{tab:test-time-scaling} indicate that the model achieves better performance through test-time scaling compared to generating directly. Additionally, the model's performance with test-time scaling improves after iterations from $M_1$ to $M_4$, which corresponds to the uptrend of the model's mathematical abilities and LLM-as-a-Judge capabilities.

\section{Conclusion}\label{sec:conclusion}
We propose a novel paradigm, Process-based Self-Rewarding Language Models, that enables LLMs to perform step-by-step long-thought mathematical reasoning and step-wise LLM-as-a-Judge simultaneously.
Given the characteristics of complex math reasoning tasks, we introduce the step-by-step reasoning, step-wise LLM-as-a-Judge and step-wise preference optimization technique into the framework.
Our results indicate that Process-based Self-Rewarding algorithm outperforms the original Self-Rewarding on a variety of complex mathematical reasoning tasks, showing potential of stronger reasoning ability better than human in the future.


\section{Limitations}\label{sec:limitations}
We aim to draw more attention to the study of adapting the self-rewarding paradigm to the complex mathematical reasoning tasks, which allows for the possibility of continual improvement beyond the human preferences. Although our new Process-based Self-Rewarding algorithm has shown effective improvements across different mathematical reasoning tasks, there are still some limitations waiting for further research. Although we successfully enable the model to perform effective step-wise LLM-as-a-Judge with a small amount of EFT data, the basic capabilities of initialized $M_1$ model directly influence the effectiveness of subsequent process-based self-rewarding. Utilizing more high-quality data to initialize LLMs more adequately may lead to stronger performance.

Additionally, due to the limited resources, we only conduct the process-based self-rewarding experiments from $M_1$ to $M_4$. Building on this, conducting experiments with more iterations to explore the impact of iteration count on LLMs' performance can help us better understand and utilize the process-based self-rewarding method.

\bibliography{custom}

\newpage

\appendix

\section{Step Number and Step Length Statistics}\label{appendix:step-stat}
We report the full results  of step number and step length across all benchmarks on the 7B and 72B models here. The 7B results are reported in Table \ref{tab:7b-step-stat}. And the 72B results are reported in Table \ref{tab:72b-step-stat}.

\begin{table*}[htbp]
    \centering
    \small
    \begin{tabular}{lcc}
        \toprule
        \textbf{Judge Strategy} & \textbf{Consistency} & \textbf{Agreement} \\
        \midrule
        Step-wise Pairwise Comparison & 0.84 & 0.88 \\
        Solution Scoring & 0.72 & 0.32 \\
        \bottomrule
    \end{tabular}
    \caption{The consistency and agreement with human evaluation of step-wise pairwise comparison and solution scoring.}
    \label{tab:solutionscore-vs-stepcompare}
\end{table*}

\begin{table*}[tbp]
    \centering
    \small
    \begin{tabular}{lccccccc}
        \toprule
        \textbf{Step Num} & \textbf{GSM8k} & \textbf{MATH} & \textbf{Gaokao2023En} & \textbf{OlympiadBench} & \textbf{AIME2024} & \textbf{AMC2023} \\
        \midrule
        $M_1$ & 5.91 & 9.35 & 8.68 & 11.75 & 7.97 & 11.18 \\
        $M_2$ & 5.24 & 8.03 & 7.43 & 9.54 & 7.03 & 9.85 \\
        $M_3$ & 4.50 & 6.43 & 5.84 & 7.36 & 7.13 & 6.9 \\
        $M_4$ & 4.09 & 5.21 & 5.11 & 6.14 & 6.4 & 5.53 \\
        \bottomrule
        \toprule
        \textbf{Step Length} & \textbf{GSM8k} & \textbf{MATH} & \textbf{Gaokao2023En} & \textbf{OlympiadBench} & \textbf{AIME2024} & \textbf{AMC2023} \\
        \midrule
        $M_1$ & 48.59 & 61.61 & 69.74 & 103.95 & 100.43 & 76.13 \\
        $M_2$ & 54.02 & 70.04 & 85.26 & 108.26 & 114.27 & 115.29 \\
        $M_3$ & 63.36 & 89.68 & 99.59 & 127.97 & 118.67 & 109.45 \\
        $M_4$ & 73.64 & 113.14 & 118.02 & 142.69 & 138.18 & 127.18 \\
        \bottomrule
    \end{tabular}
    \caption{Statistics of step number and step length on different methematical benchmarks based on 7B models.}
    \label{tab:7b-step-stat}
\end{table*}

\begin{table*}[tbp]
    \centering
    \small
    \begin{tabular}{lccccccc}
        \toprule
        \textbf{Step Num} & \textbf{GSM8k} & \textbf{MATH} & \textbf{Gaokao2023En} & \textbf{OlympiadBench} & \textbf{AIME2024} & \textbf{AMC2023} \\
        \midrule
        $M_1$ & 5.89 & 8.41 & 8.34 & 10.21 & 8.23 & 9.95 \\
        $M_2$ & 5.55 & 7.64 & 7.34 & 9.05 & 7.37 & 9.75 \\
        $M_3$ & 5.10 & 6.30 & 5.99 & 6.54 & 7.07 & 6.55 \\
        $M_4$ & 4.87 & 5.54 & 5.36 & 5.75 & 6.33 & 6.1 \\
        \bottomrule
        \toprule
        \textbf{Step Length} & \textbf{GSM8k} & \textbf{MATH} & \textbf{Gaokao2023En} & \textbf{OlympiadBench} & \textbf{AIME2024} & \textbf{AMC2023} \\
        \midrule
        $M_1$ & 47.79 & 61.00 & 69.72 & 95.38 & 104.97 & 79.36 \\
        $M_2$ & 51.19 & 67.17 & 78.00 & 101.93 & 118.08 & 86.88 \\
        $M_3$ & 57.75 & 80.46 & 91.21 & 122.53 & 118.61 & 108.95 \\
        $M_4$ & 62.86 & 96.63 & 106.28 & 134.62 & 133.66 & 113.60 \\
        \bottomrule
    \end{tabular}
    \caption{Statistics of step number and step length on different methematical benchmarks based on 72B models.}
    \label{tab:72b-step-stat}
\end{table*}

\section{Solution Scoring v.s. Step-wise Pairwise Comparison}\label{appendix:solutionscore-vs-stepcompare}

We evaluate the GPT-4o's~\citep{hurst2024gpt} consistency and agreement with humans on two different LLM-as-a-Judge strategies for complex mathematical reasoning tasks, including assigning scores to the answers and performing pairwise comparison between two individual reasoning steps. We report the results in Table \ref{tab:solutionscore-vs-stepcompare}. Our results indicate that for the complex mathematical reasoning task, step-wise pairwise comparison has better consistency and agreement with humans than solution scoring. It is highly challenging for LLMs to assign a proper and steady score to a complex long-thought multi-step solution.

\section{Prompt Templates}\label{appendix:prompt-templates}
\begin{figure*}[tbp]
    \centering
    \includegraphics[width=1.0\linewidth]{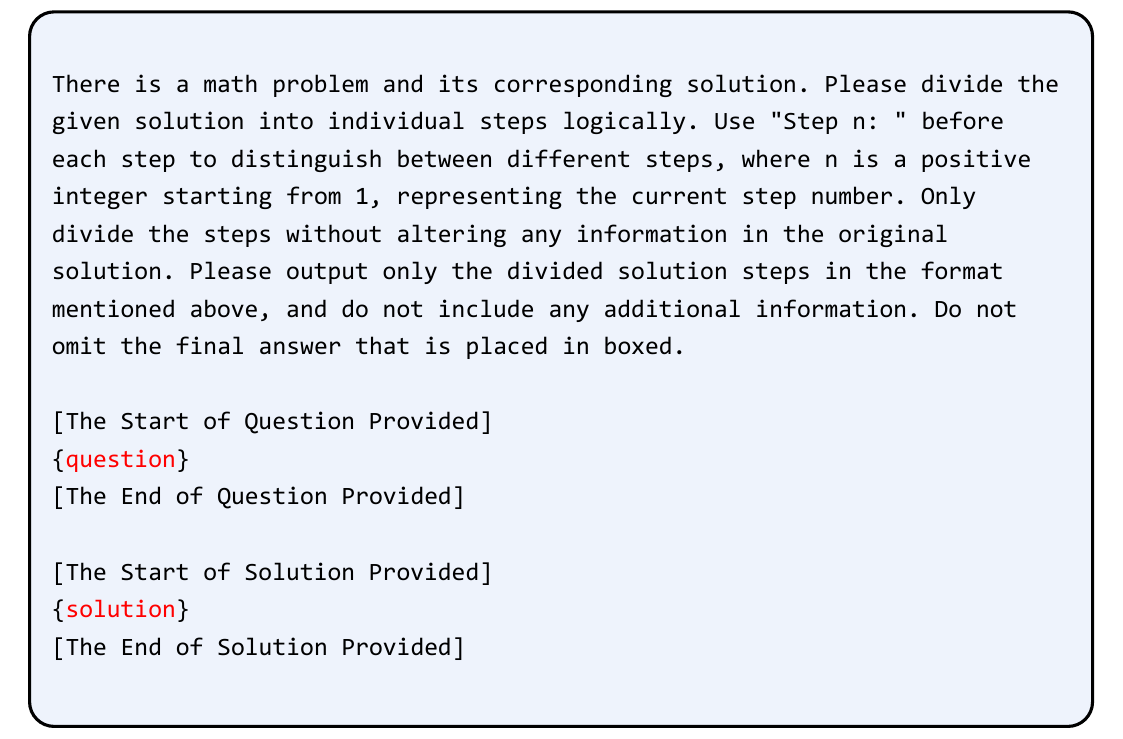}
\caption{The prompt for converting the the given solution into step-by-step format logically without altering any information in the original solution.}
\label{fig:construct-steps-prompt}
\end{figure*}

\begin{figure*}[tbp]
    \centering
    \includegraphics[width=1.0\linewidth]{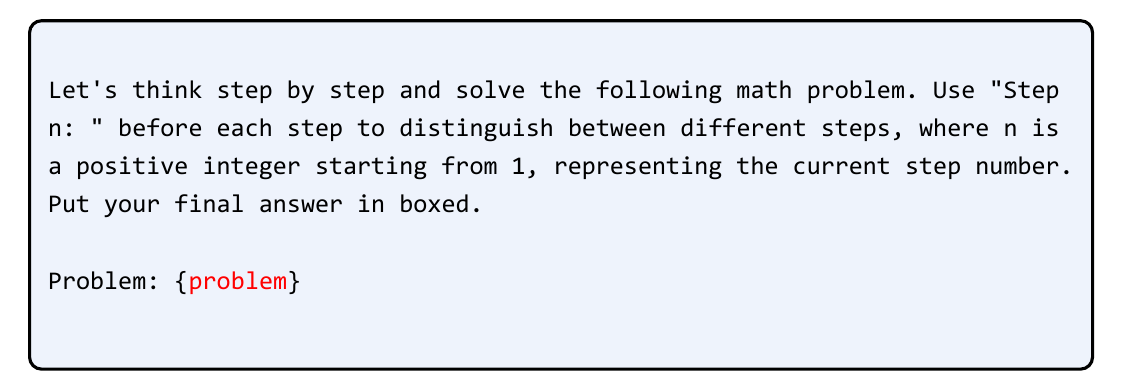}
\caption{The prompt for LLMs conducting step-by-step long-thought reasoning.}
\label{fig:math-prompt}
\end{figure*}

\begin{figure*}[tbp]
    \centering
    \includegraphics[width=1.0\linewidth]{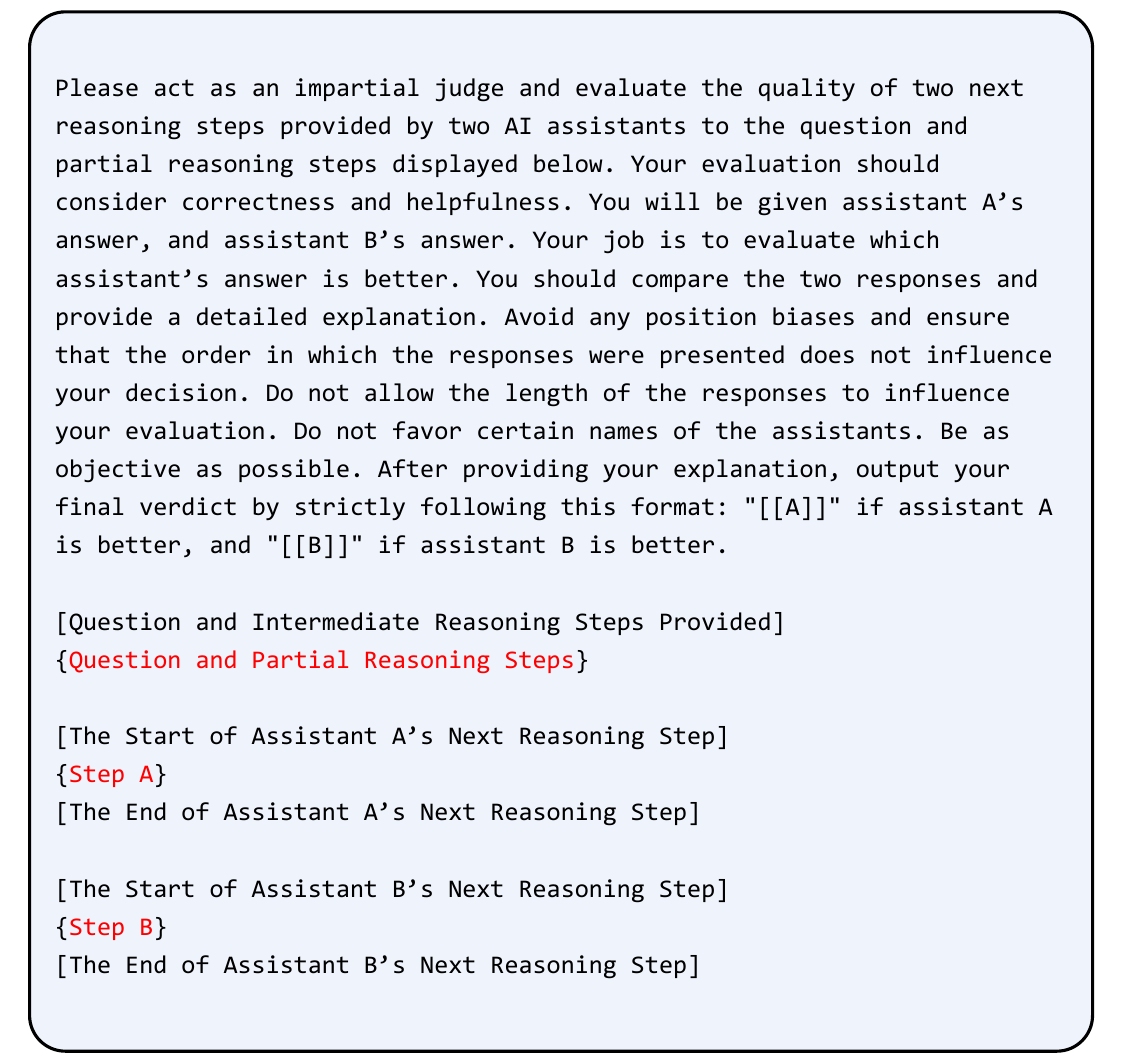}
\caption{The prompt for LLMs conducting step-wise LLM-as-a-Judge. We create this prompt template following the basic pattern of \citet{zheng2023judging}.}
\label{fig:judge-prompt}
\end{figure*}

We list the prompt templates we used in our work here. The prompt we use for constructing step-by-step formatted reasoning is shown in Figure \ref{fig:construct-steps-prompt}. And the prompts we used for step-by-step long-thought mathematical reasoning and step-wise LLM-as-a-Judge are shown in Figure \ref{fig:math-prompt} and Figure \ref{fig:judge-prompt} respectively.

\begin{table*}[tbp]
    \centering
    \small
    \begin{tabular}{lccccccc}
        \toprule
        \textbf{Setting} & \textbf{GSM8k} & \textbf{MATH} & \textbf{Gaokao2023En} & \textbf{OlympiadBench} & \textbf{AIME2024} & \textbf{AMC2023} \\
        \midrule
        $M_1$ Greedy Search & 92.6 & 76.0 & 66.2 & 41.8 & 13.3 & 45.0 \\
        $M_4$ Greedy Search & 93.7 & 76.6 & 68.1 & 44.1 & 23.3 & 57.5 \\
        $M_1$ Test-time Scaling & 94.5 & 79.1 & 64.9 & 41.6 & 16.7 & 52.5 \\
        $M_4$ Test-time Scaling & 94.5 & 79.3 & 68.3 & 43.7 & 23.3 & 65.0\\
        \bottomrule
    \end{tabular}
    \caption{The full results of greedy search and test-time scaling on 72B model.}
    \label{tab:test-time-scaling-full}
\end{table*}

\end{document}